\documentclass[10pt,twocolumn,letterpaper]{article}

\usepackage[pagenumbers]{wacv} 

\usepackage{graphicx}
\usepackage{amsmath}
\usepackage{amssymb}
\usepackage{booktabs}
\usepackage{multirow}
\usepackage{caption}
\usepackage{subcaption}
\usepackage{xcolor}
\usepackage{balance}

\usepackage[pagebackref,breaklinks,colorlinks]{hyperref}

\usepackage[capitalize]{cleveref}
\crefname{section}{Sec.}{Secs.}
\Crefname{section}{Section}{Sections}
\Crefname{table}{Table}{Tables}
\crefname{table}{Tab.}{Tabs.}

\begin{document}

\title{Analyzing the Efficacy of an LLM-Only Approach for Image-based Document Question Answering}

\author{Nidhi Hegde, Sujoy Paul, Gagan Madan, Gaurav Aggarwal \\
Google Research \\
{\tt \small \{hegden, sujoyp\}@google.com}}


\maketitle

\begin{abstract}
Recent document question answering models consist of two key components: the vision encoder, which captures layout and visual elements in images, and a Large Language Model (LLM) that helps contextualize questions to the image and supplements them with external world knowledge to generate accurate answers. However, the relative contributions of the vision encoder and the language model in these tasks remain unclear. This is especially interesting given the effectiveness of instruction-tuned LLMs, which exhibit remarkable adaptability to new tasks. To this end, we explore the following aspects in this work: (1) The efficacy of an LLM-only approach on document question answering tasks (2) strategies for serializing textual information within document images and feeding it directly to an instruction-tuned LLM, thus bypassing the need for an explicit vision encoder (3) thorough quantitative analysis on the feasibility of such an approach. Our comprehensive analysis encompasses six diverse benchmark datasets, utilizing LLMs of varying scales. Our findings reveal that a strategy exclusively reliant on the LLM yields results that are on par with or closely approach state-of-the-art performance across a range of datasets. We posit that this evaluation framework will serve as a guiding resource for selecting appropriate datasets for future research endeavors that emphasize the fundamental importance of layout and image content information.\\

\end{abstract}

\section{Introduction}


Document Question Answering (DQA) \cite{kim2022ocrfree, mathew2021docvqa} is the task of answering questions on images like receipts, papers, forms, charts, or even natural images with textual information in the scene. This essentially requires the model to develop an understanding of multiple modalities: (1) extracting useful features from raw images (2) extracting layout information (3) combining it with world knowledge to answer questions. A lot of recent vision-language models focus on training large end-to-end models from scratch \cite{lee2022pix2struct, udop, biten2021latr, huang2022layoutlmv3} aiming to utilize all the underlying modalities. 

\begin{figure}
    \centering
     \begin{subfigure}[t]{0.49\textwidth}
         \centering
         \includegraphics[height=4.2cm]{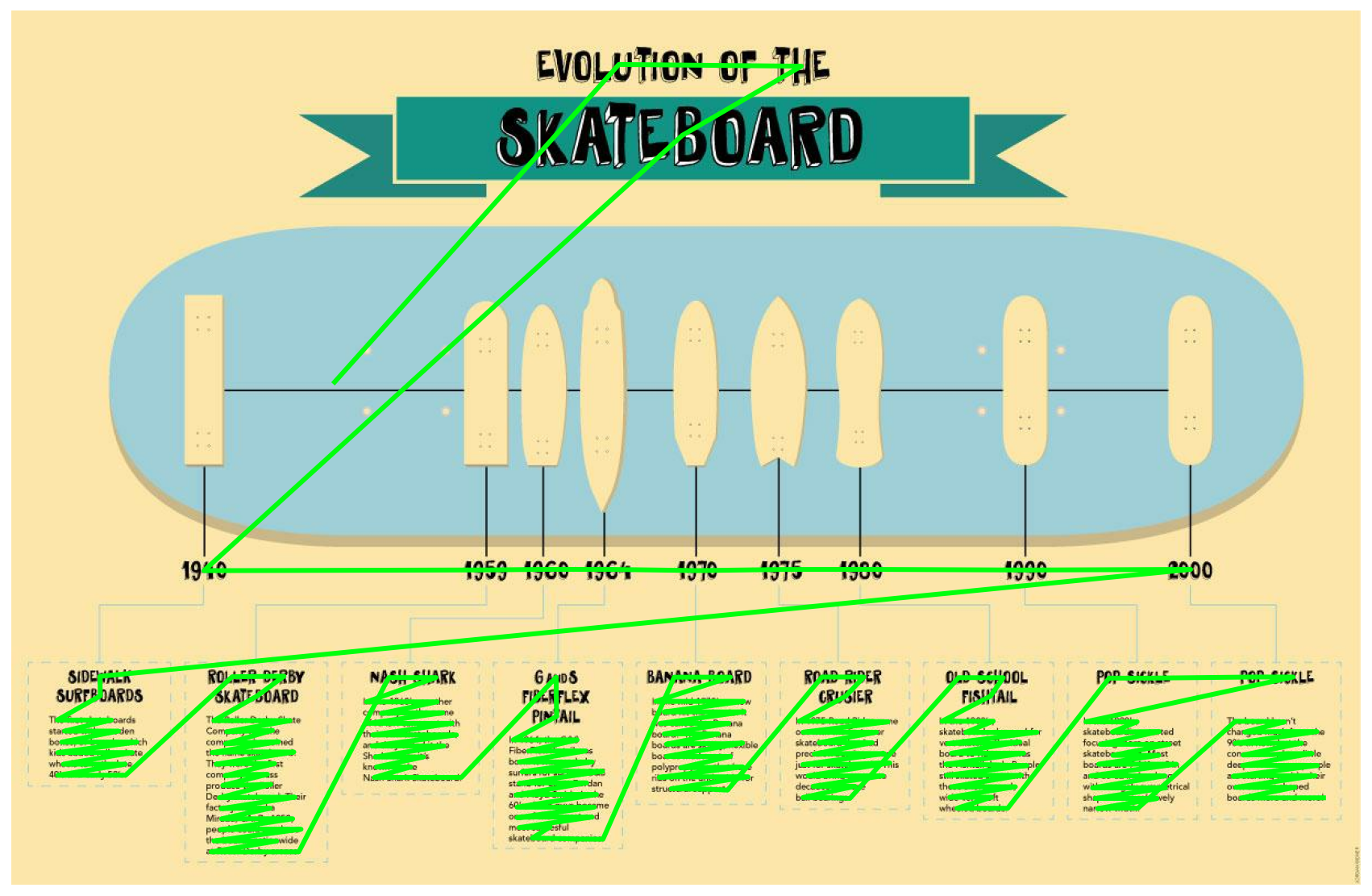}
         \caption{}
         \label{fig:ro}
     \end{subfigure}
     \begin{subfigure}[t]{0.49\textwidth}
        \centering
        \includegraphics[height=4.2cm]{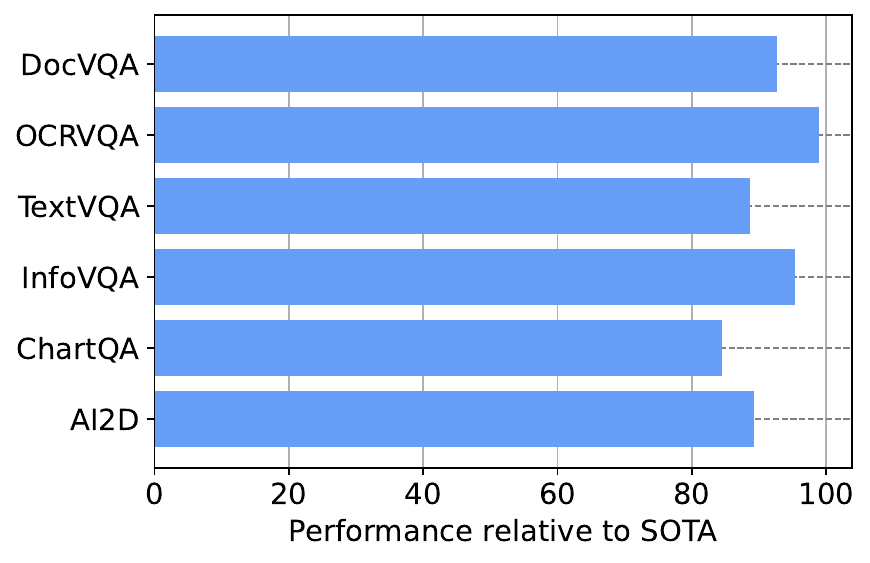}
        \caption{}
        \label{fig:rel_perf}
     \end{subfigure}
    \caption{(a) \textbf{Reading Order} is an interesting way of serializing text information. An example of reading order can be observed by following the trajectory of the green line. (b) Analyzing the performance w.r.t. SOTA of a language-only model with only reading ordered text input.}
\end{figure}


These models typically consist of three kinds of architectures: encoder-decoder \cite{t5, lewis2019bart}, encoder-only \cite{devlin2019bert} and decoder-only \cite{chowdhery2022palm, brown2020language}. The input to these models, i.e., the image and the question, is processed in three ways - (1) raw image features are extracted using 2D patch based tokens retaining the layout information (2)  text is detected using optical character recognition (OCR) \cite{ocr_2, ocr_3}, and then encoded as tokens with spatial position information (3) the question is decomposed into tokens using purely language based tokenizers. These networks are typically pre-trained individually or end-to-end with self-supervised or weakly-supervised losses, and then finally finetuned for the end task of DQA on labeled training data.


While recent advancements in document-image question answering have relied on combining multiple modalities, the impact of the individual components is unclear. For example, it is difficult to say whether a newly proposed method improves due to a better vision encoder, a stronger language model with better reasoning capabilities, or both. In some of these tasks, the language model's knowledge may be enough to answer a question, even without the image content or layout, while in others, the image content and/or layout could be truly useful. In this work, we analyse the task of document image question answering from an LLM-only perspective, without encoding the visual components or any layout information.


As humans, we read text entities in images serially, then use the layout information to correlate these entities, constructing a readable sequence of text in a specific order, called the "reading order". An example of reading order is shown in Figure \ref{fig:ro}. Recent advances in reading order prediction on documents \cite{wang2023text, wang2021layoutreader} provide a way to obtain this serialized text from 2D images. This text can be fed to text-only large language models, such as Flan models \cite{chung2022scaling}, which have been instruction-tuned to solve text-only question-answering tasks. Such a setup does not involve any vision encoder or position information other than the reading order information. We show that this approach can perform well, coming close to the state-of-the-art methods that explicitly encode visual information (Figure \ref{fig:rel_perf}). This suggests that language models can be effective for DQA, even when access to visual information is limited.

In this work, we study the challenging problem of DQA by viewing it purely as a language modeling task. Specifically, we explore how well can pre-trained LLM's answer questions about images, using only the text in the image serialized using the reading order.
We perform ablation studies, qualitative and quantitative analysis to understand when text-only language models work well and when they need more information from a vision encoder. Our results suggest that text-only language models can be effective for DQA, but they may need more information for certain types of tasks. Our work can serve as a good yardstick for future research on DQA, to decide whether a task can be solved purely through text-based methods or if it needs additional image and layout information. Note that the goal of this work is not propose an algorithm without a vision encoder that works well for all tasks and question types, nor is it to reduce the computation cost of these visual-language models (VLM's). Rather the goal is to analyze how well does tokenizing the textual information in an image using only reading order along with pre-trained language models, helps in solving question answering tasks.



Overall, our contributions are as follows: 

\begin{enumerate}
    \item We model document question answering as a pure language modeling task, with images tokenized to text using reading order.
    \item We study how well language models can perform on document question answering tasks without using a vision encoder.
    \item We perform quantitative analysis correlating the performance of the proposed model with factors, like the reading order perplexity, task lengths, answer contents, \etc, to identify the principles guiding the success or failure of language-only models.
\end{enumerate}

\section{Related Work}
We categorize the prior works in this domain into two categories - OCR-based methods and OCR-free methods. 

\textbf{OCR-based methods} typically entail two forms of input - the raw image and text tokens, which are obtained via Optical Character Recognition (OCR) along with their respective positions. Models such as LayoutLM \cite{huang2022layoutlmv3}, LaTR \cite{biten2021latr}, FormNet \cite{lee2022formnet}, DocFormer \cite{appalaraju2021docformer}, UDOP \cite{udop}, and M4C \cite{hu2020iterative} are classified within this category. Methods, such as M4C, also incorporate additional inputs like detected objects. Notably, OCR-Based Methods encompass strategies for understanding both scene-text and documents. The majority of these methods employ complex techniques to fuse text and image tokens in order to enhance learning capabilities. For instance, in the UDOP model \cite{udop}, each detected text token is coupled with the corresponding image token prior to being sent to the transformer encoder. The FormNet model \cite{lee2022formnet} introduces an extra GCN layer to the input text tokens before transmitting them to the transformer layer.

\textbf{OCR-Free Methods}: These strategies do not utilize any explicit OCR data, instead, they solely depend on the visual information in the image and let the network learn the text information implicitly. Often they build over a pre-training step that makes it easy for the network to learn text information. The DONUT model \cite{kim2022ocrfree} presents an OCR-free Visual Document Understanding (VDU) model which is grounded in a transformer architecture. It employs reading order prediction as a pre-training task. The Pix2Struct model \cite{lee2022pix2struct} incorporates the question as part of the image by appending it to the image header. As a pre-training objective, it learns to predict HTML which can render the image. Despite its high performance on various downstream document understanding tasks, it falls short on tasks that are text-heavy. The PaLI model \cite{chen2022pali} is reliant on an encoder-decoder architecture, utilizing a ViT \cite{dosovitskiy2020image} based encoder to encode the visual components in the image. It is also trained to predict text tokens in the image. A drawback of OCR-free approaches is that they usually require the vision encoder to operate at a significantly higher resolution (to be able to "read" the text), which can substantially increase the computational cost.

In summary, contemporary document understanding methods either rely solely on visual data or a combination of visual features and text (OCR). In our work, we explore the capabilities and limitations of text-only models to perform Visual Question Answering (VQA) on documents using solely the textual context obtained from an OCR pipeline, given in a specific order.

\begin{figure*}
    \centering
    \includegraphics[scale=0.65]{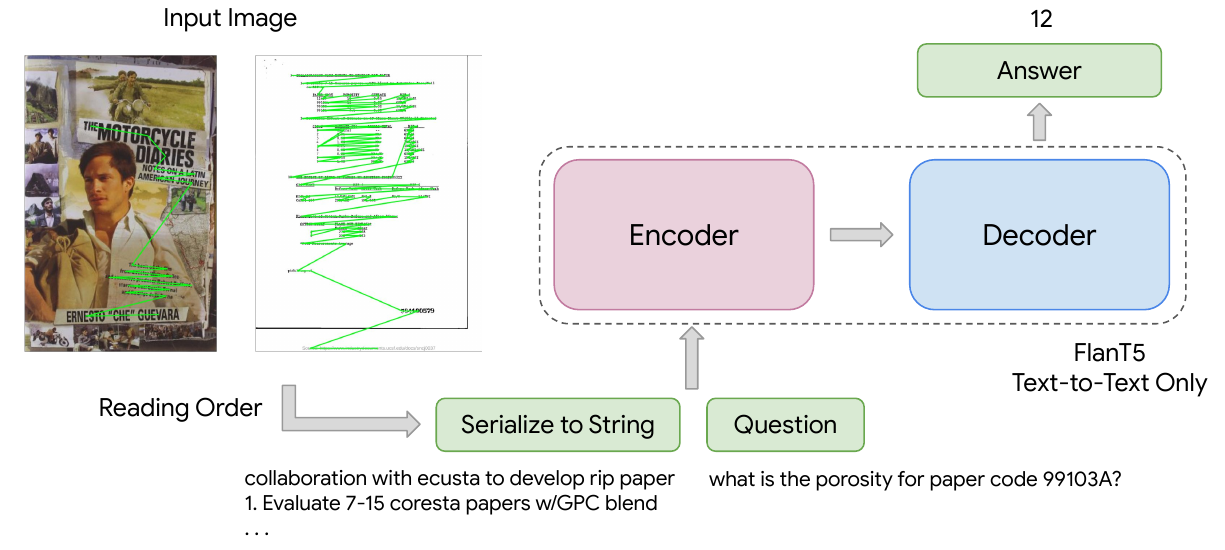}
    \caption{\textbf{Using LLMs for Document Question Answering on Images.} We first extract text from the image using an off-the-shelf OCR engine along with the reading order \cite{wang2023text}, i.e., the order a human is likely to read the text in the image. This is then serialized to a string, and sent to the encoder along with the question. The answer is then obtained from the output of the decoder. Note that we use FlanT5 \cite{chung2022scaling} text-to-text model, which has been instruction tuned on a variety of tasks. The green line segments on the input images denote the reading order into which the text in the image is serialized and fed to the encoder.}
    \label{fig:framework}
\end{figure*}

\section{Modeling DocumentQA as an LLM-only Task}
In this section, we discuss the design of our experimental setup: the process of tokenizing 2D information into a 1D token stream, and converting a multi-modal task into a text-only language modeling task.

{\flushleft \textbf{Background.}} 
Typical DocumentQA frameworks capture layout information in document images by either tokenizing raw images into patches using a ViT-style architecture and/or detecting words using OCR, which are then sent as tokens with spatial position embeddings. These two tokenization schemes have proven effective for solving document QA tasks. However, our primary focus in this paper is to evaluate the performance of a large language model in solving the task without using any image or absolute position information.

{\flushleft \textbf{Reading Order.}} 
The reading order of text in documents, refers to the organization of text in an image into a coherent and meaningful 1-dimensional sequence. Optical character recognition (OCR) engines typically identify regions of text and determine the correct input text order based on the layout analysis. The challenge of detecting the correct reading order varies greatly depending on the complexity of the input text layout~\cite{wang2023text}.
Existing reading order extraction models \cite{wang2021layoutreader, clausner2013significance} in the literature  are generally language agnostic, relying on layout information rather than content features to predict the order. These models are trained on curated data, such as books where the reading order is well-defined, or manually annotated data from in-the-wild images. 

In our work, we use reading order to convert data in 2D images into a token stream by concatenating the text. Although the reading order may not be perfect in complex documents, our analysis demonstrates that this method works reasonably well for a large number of datasets used in the literature. Moreover, we show that domain knowledge about the dataset can be incorporated to improve the reading order using rule-based heuristics, leading to better QA task performance.

\paragraph{Visual QA as a text-based task.} 
We design document image question-answering as a purely language modeling task. We obtain the text in the reading order using \cite{wang2023text} and concatenate them to form a string $\mathcal{C}$, which is then sent to a large language model for QA. We refer to this textual representation as the "OCR context" throughout the paper. Using this context, we prompt the large-language model with a template that includes the OCR context, question, and a placeholder for the answer:
\begin{equation}
    \text{Context: } \mathcal{C}, \text{Question: } \mathcal{Q}, \text{Answer: } \_\_\_\_
    \label{eqn:prompt}
\end{equation}
This prompt is generic enough to be used with either an encoder-decoder or a decoder only class of LLM's. But owing to the success of Flan-T5 family of models and its open-source availability, we use it for all experiments in this paper. Flan-T5 \cite{chung2022scaling} is an encoder-decoder model obtained by instruction tuning of T5 \cite{t5} which has been shown to be effective on a wide range of NLP tasks, including text summarization, question answering, code generation, and many more. The prompt in Eqn \ref{eqn:prompt} goes as input to the encoder of the model, and the output is then obtained from the decoder. Figure \ref{fig:framework} shows the flow of the model from input image to output text. 

We observe that extracting reading order is non-trivial, and the performance of text-only models to address document QA tasks depends significantly on it.
Interestingly, we find that our scheme offers reasonable zero-shot performance without any finetuning. After finetuning, the performance improves significantly as the network learns to have its predictions more aligned with the data, as well as to reason from irregular text arising from the incorrectness of the reading order.

\section{Experiments}

\subsection{Datasets} \label{sec:datasets}
We evaluate the performance of this method on six benchmark datasets, which span multiple domains: documents, infographics, tabular data, illustrative science diagrams, book covers and scene text. The details of all the datasets used in our experiments are provided in Table \ref{table:datasets}.
\begin{table}[t]
			\caption{Details of datasets used in this paper for experiments.}
			\label{table:datasets}
			\centering
			\renewcommand{\arraystretch}{1.2}
	        \setlength{\tabcolsep}{4pt}
	        \resizebox{0.48\textwidth}{!}{
			\begin{tabular}{l c c c | c c c c}
		    \toprule
		    & \multicolumn{3}{c|}{No. of samples} & \multicolumn{3}{c}{Median no. of characters} \\
		    &  Train & Val & Test & Text & Question & Answer \\
		    \midrule
		    OCR-VQA \cite{mishra2019ocr} & 801k & 100k & 100k &  115 &  31 & 12 \\
		    DocVQA \cite{mathew2021docvqa} &  39.5k &  5.3k & 5.1k &  1055 & 41 & 10 \\
		    InfoVQA \cite{mathew2022infographicvqa} &  23.9k &  2.8k &  3.2k &  1567 & 65 & 5 \\
		     TextVQA \cite{singh2019towards} & 34.6k & 5k & 5.7k & 71 & 33 & 7 \\
		     ChartQA \cite{masry2022chartqa} & 28.3k & 1.9k & 2.5k & 279 & 64 & 4\\
		      AI2D \cite{kembhavi2016diagram} & 12.3k & 120 & 3k & 100 & 47 & 7\\
		   \bottomrule
	    \end{tabular}
	    }
\end{table}

{\flushleft \textbf{OCR-VQA} \cite{mishra2019ocr}} is a dataset of book covers, with questions based on metadata from the book such as title, author, genre, etc. The metric used for this dataset is exact match accuracy.

{\flushleft \textbf{DocVQA} \cite{mathew2021docvqa}} is a dataset of scanned documents with printed and type-written text in various layouts, with/without natural images in them. Some images also contain handwritten data. The metric used for this dataset is Average Normalized Levenshtein Score (ANLS) \cite{biten2019icdar}. 

{\flushleft \textbf{TextVQA} \cite{singh2019towards}} consists of scene-text images from Open Images v3 with questions that require reading the text to answer. The metric used for this dataset is VQA accuracy.

{\flushleft \textbf{InfographicsVQA} \cite{mathew2022infographicvqa}} is a dataset of infographic documents that include text, visual and graphical information with questions from tables, figures and visualizations. Questions often involve combining multiple cues, making this one of the more challenging datasets for document understanding. The metric used for this dataset is Average Normalized Levenshtein Score (ANLS).

{\flushleft \textbf{ChartQA} \cite{masry2022chartqa}} is a dataset of questions on visual representations of tabular data (like bar charts, pie diagrams, \etc). The metric used for this dataset is relaxed accuracy, i.e., exact match but tolerating 5\% of numerical error.

{\flushleft \textbf{AI2D} \cite{kembhavi2016diagram}} consists of multiple choice questions on illustrative diagrams spanning various scientific domains. The metric used for this dataset is exact match accuracy.

\begin{table*}
  \caption{\textbf{Zero-shot} performance of FlanT5 models for document QA with reading order input}
  \label{table:zeroshot}
  \centering
  \resizebox{0.75\textwidth}{!}{%
    \begin{tabular}{@{}lccccccc@{}}
    \toprule
    Model            & Params & OCR-VQA \cite{mishra2019ocr} & DocVQA \cite{mathew2021docvqa} & InfoVQA \cite{mathew2022infographicvqa} & TextVQA \cite{singh2019towards} & ChartQA \cite{masry2022chartqa} & AI2D \cite{kembhavi2016diagram}\\ \midrule
    FlanT5-B      &     250M   &   34.2      &  33.6      &    14.1             &   32.5    & 10.5 & 22.8  \\
    FlanT5-L     &   780M     &     36.0    &   35.1     &         21.3      &       36.7  & 12.1 & 24.0 \\
    FlanT5-XL        &   3B     &   38.3      & 48.4       &        27.3         &    42.3   & 14.8 & 41.3  \\
    FlanT5-XXL       &     11B   &     42.4      &     54.1   &       34.0          &     44.0  & 20.5 & 50.8 \\
    \midrule
    {\color{gray}
    \multirow{2}{*}{SOTA (Finetuned)}} 
     & & {\color{gray} 71.3} & {\color{gray} 84.7} & {\color{gray} 47.4} & {\color{gray} 71.8} & {\color{gray} 64.2 } & {\color{gray} 76.2}\\
     & & {\color{gray} \tiny (Pix2Struct)} & {\color{gray} \tiny (UDOP)} & {\color{gray} \tiny (UDOP)} & {\color{gray} \tiny (PaLI)}  & {\color{gray} \tiny (MATCHA)}  & {\color{gray} \tiny (DisAVR)}  \\

    \bottomrule
    \end{tabular}
    }
\end{table*}
\begin{table*}
  \caption{\textbf{Finetuned} results on six datasets and comparisons with the state-of-the-art results from literature. All the methods in the top portion use a vision encoder and/or position information of the text in the images, whereas the bottom section only use a sequence of reading ordered text as input. The best results in each group are highlighted in bold.}
  
  \label{table:finetune}
  \centering
  \resizebox{0.75\textwidth}{!}
  {
    \begin{tabular}{@{}clccccccc@{}}
    \toprule
    & Model            &  OCR-VQA \cite{mishra2019ocr} & DocVQA \cite{mathew2021docvqa} & InfoVQA \cite{mathew2022infographicvqa} & TextVQA \cite{singh2019towards} & ChartQA \cite{masry2022chartqa} & AI2D \cite{kembhavi2016diagram} \\ 
    \midrule
    \multirow{8}{*}{\rotatebox{90}{\small Vision Encoder}} & LaTR-B \cite{biten2021latr}   &        67.5    &     -   &      -       &       59.5  &   - & -  \\
    & LaTR-L  \cite{biten2021latr}     &    -     &     -   &       -          &        61.1 &   - & -   \\
    & Layout LMv3 \cite{huang2022layoutlmv3}     &    -     &     83.4  &            -     &   -    &   - & -  \\
    & DONUT  \cite{kim2022ocrfree}   & - & 67.5 & - & - &   - & -  \\

    & Pix2Struct-B \cite{lee2022pix2struct}  &     69.4    &    72.1    &        38.2         &   -     &   56.0 & 40.9  \\
    & Pix2Struct-L \cite{lee2022pix2struct}    &      \textbf{71.3}   &     76.6   &         40.0       &    -    &   58.6 & 42.1  \\ 
    & UDOP  \cite{udop}              &    -     &  \textbf{84.7}      &        \textbf{47.4}         &     -   &   - & -   \\
    & PaLI-3B \cite{chen2022pali}     &    -     &     -   &            -     & 60.1 &   - & -     \\ 
    & PaLI-17B \cite{chen2022pali}     &    -     &     -   &            -     & \textbf{71.8} &   - & -     \\ 
    
    & MATCHA \cite{liu2023matcha} & 68.9 & 74.2 & 37.2 & - & \textbf{64.2} & 42.6\\
    & DisAVR \cite{disavr}     &    -     &     -   &            -     &    - & - &  \textbf{76.2}    \\ 
    \midrule
    \multirow{4}{*}{\rotatebox{90}{\small Only LLM}} & FlanT5-B (250M)      &     67.2   &    69.2    &      31.7           &    54.5    & 45.8 & 58.7\\
    & FlanT5-L   (780M)    &     68.5    &   73.6    &       37.4          &     58.4    & 50.7 & 64.4\\
    & FlanT5-XL  (3B)      &     69.6    &     77.2   &         42.2        &     61.4    & 52.2 & 68.0\\
    & FlanT5-XXL   (11B)    &     \textbf{70.5}    &    \textbf{78.5}     &         \textbf{45.2}      &     \textbf{63.6} & \textbf{54.2} & \textbf{68.2} \\
    \bottomrule
    \end{tabular}
    }
\end{table*}

\subsection{Models and Implementation Details} \label{sec:impl}
We evaluate the performance of four Flan-T5 \cite{chung2022scaling} model sizes - base (B) with 250M parameters, large (L) with 780M parameters, XL with 3B parameters and XXL with 11B parameters.
The B, L and XL variants are fine-tuned for 200K steps with a batch size of $512$ and the XXL variant is fine-tuned with a batch size of $128$ for 400K steps. The input sequence length is set to $1024$ for most datasets, except for OCR-VQA, which has an input length of $128$. The target sequence length is set to $32$ for all datasets. These sequence length calculations are based on token length analysis of the respective datasets. A learning rate of $0.001$ and a weight decay of $0.1$ are used throughout the fine-tuning process. Also note that no explicit hyperparameter tuning is conducted for any of the reported results. All experiments are done in the T5X framework\cite{roberts2022scaling} on Google Cloud TPUs.
\subsection{Zero-Shot DocumentQA}
Given the Flan-T5 models have been instruction-tuned on a collection of tasks, they generally perform well on new tasks. We first analyze their zero-shot performance across all datasets and model sizes, with results shown in Table \ref{table:zeroshot}. We see a clear trend of increasing performance with larger model sizes. It is important to note that all state-of-the-art (SOTA) models have been fine-tuned using either visual or layout information, or both.
However, the zero-shot performance of FlanT5-XXL is reasonably good, considering it lacks a vision encoder and has not been fine-tuned on the dataset. For instance, we see that the zero-shot performance for InfoVQA is just $4.2\%$ behind the finetuned performance of Pix2Struct-B (refer Table \ref{table:finetune}), and is not far behind UDOP, the SOTA model, with only a $13.4 \%$ difference. Better prompting techniques than Eqn~\ref{eqn:prompt} could potentially further enhance performance, as has been the case with similar large language models (LLMs) in the literature.


\subsection{Finetuning for DocumentQA}
In this analysis, we finetune the pre-trained FlanT5 models on the training set of the individual datasets, with details provided in Section \ref{sec:impl}. We present the results in Table \ref{table:finetune}.
As expected, the fine-tuned models yield significantly better performance than zero-shot performance. In some cases such as OCR-VQA and InfoVQA, the finetuned model performs very close to the SOTA methods. For OCR-VQA, it reaches a performance of $70.5 \%$, only $0.8\%$ behind the SOTA, and for InfoVQA, it falls behind by $2\%$. In most cases, this LLM only approach outperforms recent baselines from the literature that utilize a vision encoder. For instance, it surpasses LaTR-L by $2.5\%$ on TextVQA and Pix2Struct-L by $5.2\%$ on InfoVQA, MATCHA by $25.6\%$ on AI2D, and so on. This suggests that a lot of questions in these datasets can be answered by an LLM-only model, without any visual inputs or detailed layout information. Hence, a strong language model can do a good job in understanding the question, retrieving information from world knowledge and combining it with the context to come up with an answer.

We also observe a consistent performance improvement as we go from FlanT5-B to FlanT5-XXL which suggests that larger models tend to perform better for such tasks. It is also interesting to note that for an easier task, such as OCR-VQA, the performance improvement we observe by scaling the language model is much smaller than a complex task such as InfoVQA. Overall, the reading order + LLM scheme falls short only by a few percentages compared to state-of-the-art methods in literature, all of which include a vision encoder. This prompts us to investigate the factors behind the cases where an LLM-only scheme fails for DocumentQA and whether it is possible to overcome these limitations.

We identify three key factors that seem to play a key a role in determining the effectiveness of an LLM-only setup for DocumentQA tasks: (1) \textbf{Reading order quality}, (2) \textbf{OCR Context Length}, and (3) \textbf{Presence of answer in the image text}. In the subsequent sections, we examine these factors through empirical studies to demonstrate how these factors correlate with the performance.
\begin{figure}
    \centering
     \begin{subfigure}[t]{0.49\textwidth}
        \centering
        \includegraphics[height=4.2cm]{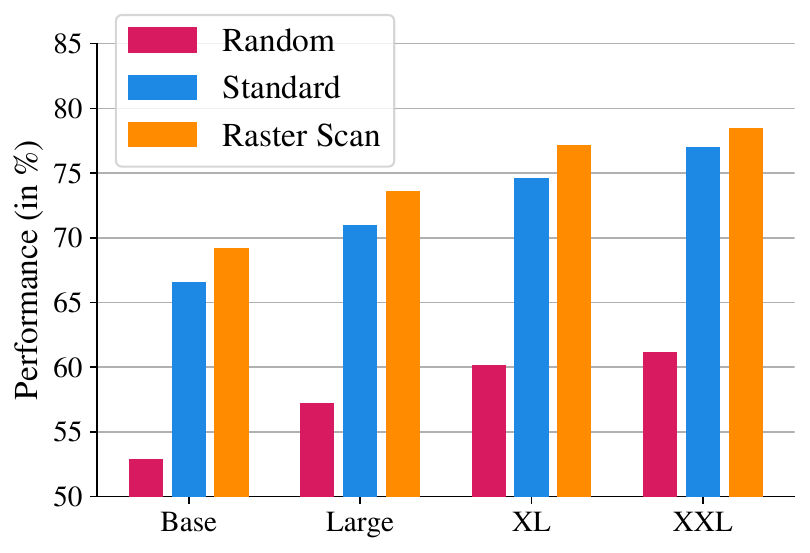}
        \caption{}
        \label{fig: ro_model_size}
     \end{subfigure}
     \begin{subfigure}[t]{0.49\textwidth}
         \centering
         \includegraphics[height=4.2cm]{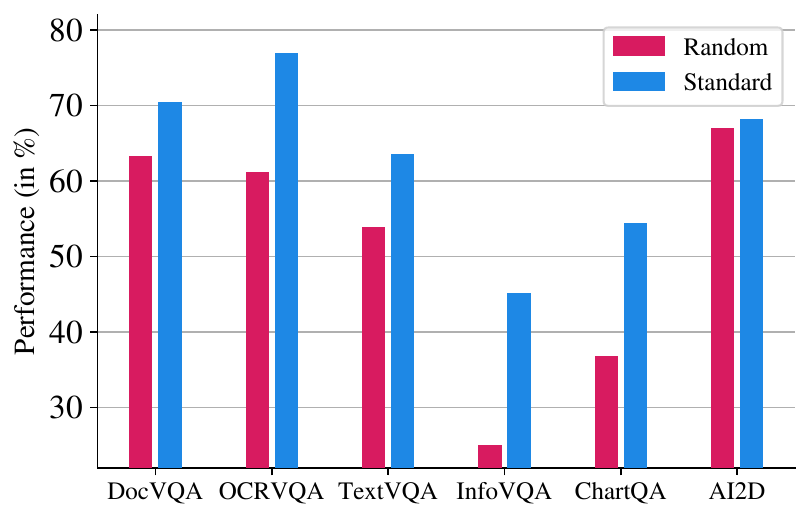}
         \caption{}
         \label{fig:ro_dataset}
     \end{subfigure}
      \caption{(a) The impact of using a better reading order \textbf{on different model sizes} for DocVQA dataset. Note that \textit{Raster Scan} is a strategy we specifically design for DocVQA which outperforms the standard reading order obtained from OCR pipeline (b) The comparison of how the choice of the reading order influences performance \textbf{across different datasets} for FlanT5 XXL. We see that InfoVQA is affected the most by shuffling the reading order. Values plotted are against the respective metrics used for the datasets}
      \label{fig: ro}
\end{figure}

\subsection{Reading Order Quality}
\begin{figure*}
    \centering
    \includegraphics[width=0.6\textwidth]{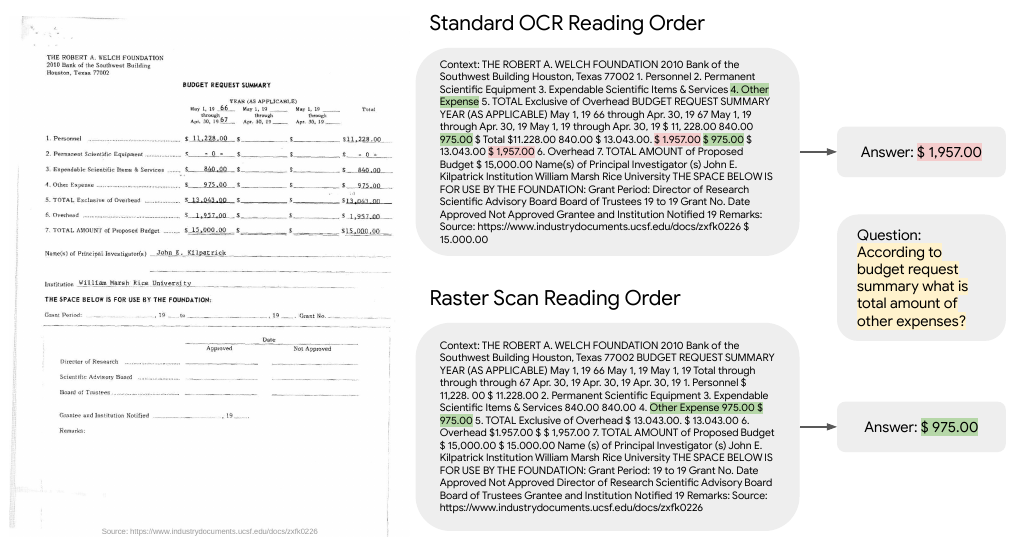}
    \caption{Text from two different reading orders along with the answer for the same question for both cases. As we can see for this example, the standard reading order is jumbled up, whereas the raster scan based reading order does a better job at extracting the information from the text, and thus enables the LLM to answer the question correctly.}
    \label{fig:raster_scan}
\end{figure*}

\begin{figure*}[t]
     \centering
     \begin{subfigure}[t]{0.32\textwidth}
         \centering
         \includegraphics[scale=0.35]{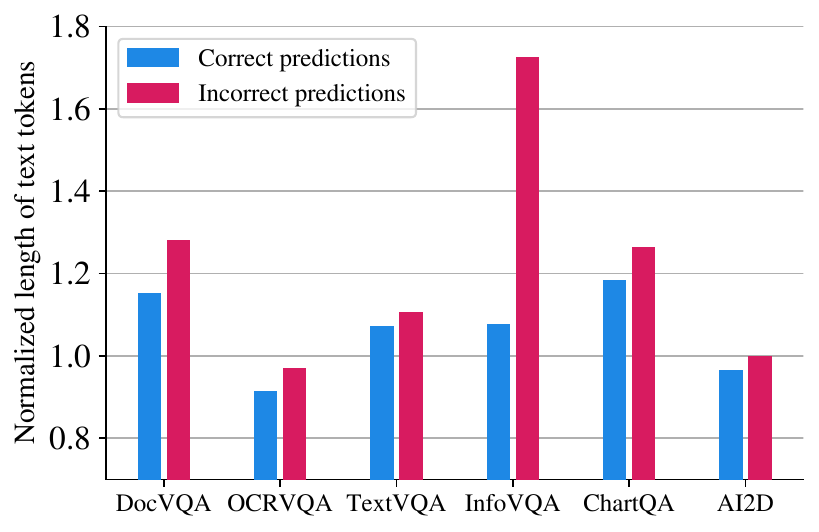}
         \caption{}
         \label{fig:len_ocr}
     \end{subfigure}
      \begin{subfigure}[t]{0.32\textwidth}
        \centering
        \includegraphics[scale=0.35]{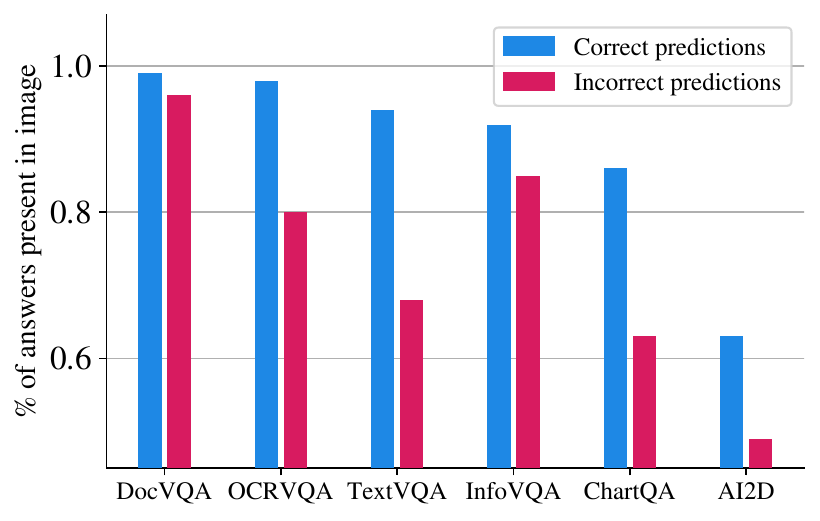}
        \caption{}
        \label{fig:percent_ans}
     \end{subfigure}
     \begin{subfigure}[t]{0.32\textwidth}
         \centering
         \includegraphics[scale=0.35]{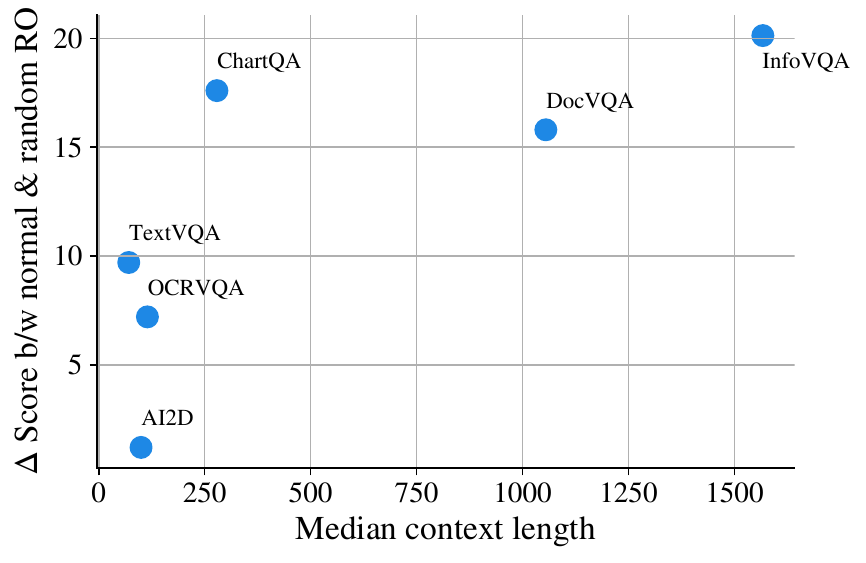}
         \caption{}
        \label{fig:context_len_vs_score}
     \end{subfigure}
     \caption{(a) \textbf{Normalized length of text in image} for the set of correct and incorrect answers.  (b) \textbf{Percentage of answers in text} for correct and incorrect set per dataset. Results are on FlanT5-B. (c) Difference in performance between normal and random reading order with median context length. The difference is much higher when context length is larger.}
\end{figure*}

A good reading order is crucial to obtain the right context $\mathcal{C}$ in the scheme outlined in Eqn \ref{eqn:prompt}. 
While it may be hard to design a universally optimal strategy to get reading orders that work well for different tasks, it is possible to design task-specific heuristics to improve a given reading order. With this in mind, we come up with a basic reading order strategy and test it on DocVQA.

The basic reading order strategy can be explained as follows: OCR provides a set of bounding boxes where the text appears in the image. We begin with the uppermost and leftmost bounding box. Then, we proceed to the next bounding box along the width, and get all the bounding boxes whose centroids fall within a specific threshold from the current one. Due to similarity of this technique to raster scanning of 2D images, we label this strategy as \textit{raster scan}. An example of the same if shown in Figure~\ref{fig:raster_scan}. 
Using this basic strategy, we observe a significant improvement in performance, indicating the importance of a proper reading order for document understanding tasks (Figure~\ref{fig: ro_model_size}). 

It is important to note that such a basic strategy may not be as effective for other kinds of documents. 
The goal of this experiment is just to demonstrate the effect of an appropriately chosen/crafted reading order to enhance performance. If the documents involved are somewhat homogeneous in nature, coming up with an appropriate reading order heuristic may not be too hard. As an ablation, we shuffle the reading order and provide the shuffled text as input to the LLM, which is then fine-tuned on the shuffled reading order text. As shown in~(Figure \ref{fig:ro_dataset}), the performance declines considerably, which is not unexpected. This further highlights the significance of reading order for the proposed scheme of pure text-based modeling without a vision encoder. For completeness, we compare different choices of reading orders across different model sizes (Figure~\ref{fig: ro_model_size}), and different datasets (Figure~\ref{fig:ro_dataset}). The standard reading order refers to \cite{wang2023text}, which is what we use for most of the experiments unless otherwise mentioned. 

We also observe that the effect of reading order importance increases significantly with larger context lengths. This is evident from Figure \ref{fig:context_len_vs_score} where we observe that the difference in performance between standard and random reading order is significantly higher for datasets with longer context lengths. From these experiments and observations, we can conclude that while layout information may be important, using a simple ordering based information (reading order) can be enough to answer a major chunk of the questions in these datasets. So ideally, to understand if an algorithm is able to extract better layout information from documents, reading order can act as a strong baseline.

\subsection{Effect of OCR Context Length}
In DocumentQA tasks, the model must derive reasoning from the content of the image, which includes both text and the image itself. As the density of text in the image increases, it becomes more challenging for the model to extract the answer. We compare the median lengths of the text tokens in images for the set of correctly and incorrectly predicted answers to empirically verify this hypothesis. We plot these statistics across different datasets in Figure \ref{fig:len_ocr}. To display the numbers on the same plot, we normalize them with the median length of tokens for the entire dataset. The results, in Figure \ref{fig:len_ocr}, indicate that the model makes more errors when the text length in the image is longer. For InfoVQA, note that the median length of incorrect answers is significantly higher than the set of correct answers in comparison to other datasets. For datasets such as TextVQA and AI2D, the difference in context length between incorrect and correct answers is not high indicating that may not be a factor behind the correctness. This is intuitive as in these datasets, there are a lot of questions which need visual understanding that goes beyond the text content of the image, such as the visual characteristics of objects in the image, their relative positions and so on. 

\subsection{What type of questions cannot be answered?}
By design, our method can only answer questions if the text content of the image contains the answer, or questions which can be answered using only world knowledge. Therefore, we develop a metric to determine the percentage of questions that can be answered using only text, i.e., what fraction of the answers lies within the serialized text for a given example. Based on whether our model correctly or incorrectly predicts the answer, we perform an averaging over the dataset to obtain 2 values: one for the correctly predicted set and the other for the incorrect set. We compare these values in Figure \ref{fig:percent_ans}. As evident from the plot, for the set of incorrect answers, the percentage of answers present in the image as text is lower than for the set of correct answers. Note that we remove all Yes/No questions before calculating these numbers. Also, for OCR-VQA, we remove the genre classification questions, since these have to be inferred and not directly extracted from the context.

\section{Discussion}
In this section, we discuss some challenges and metrics that correlate strongly with model performance for a LLM-only document image question-answering model.

\paragraph{Reading Order Perplexity.} As evident from Table \ref{table:finetune}, even without using a vision encoder, the performance is quite high. However, in some cases either we do not have enough information from the text or the reading order itself is jumbled that a language only model may find it difficult to answer the questions. Intuitively we expect the performance of any language model on a question answering task to be strongly correlated with the kind of data it has been trained on. In order to quantify this in the form of a heuristic, we define a metric called \textit{Reading Order Perplexity}.

\begin{figure*}[!htb]
    \centering
    \includegraphics[scale=0.32]{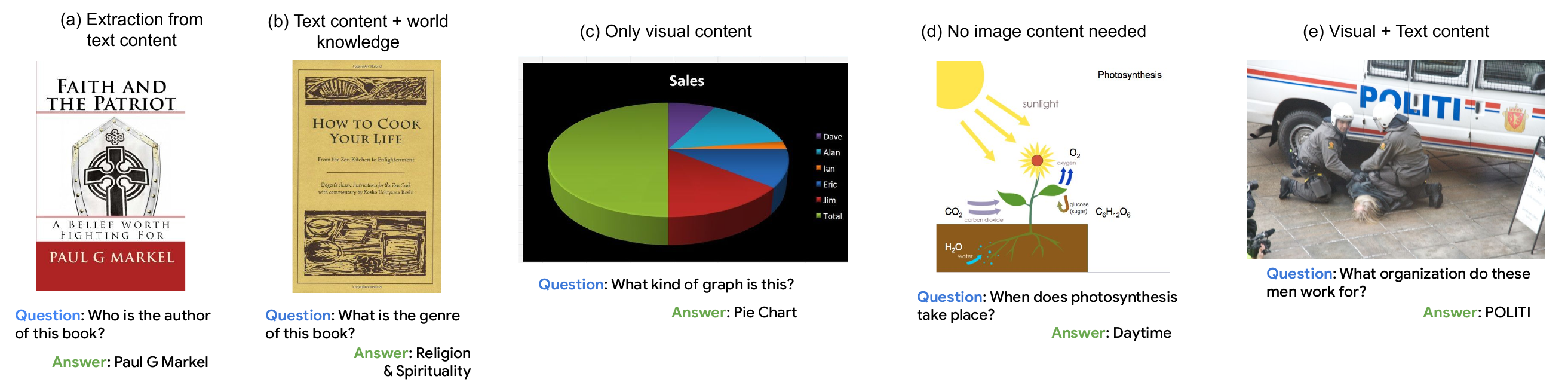}
    \caption{Example document image question-answer pairs for different types of questions.}
    \label{fig:exemplars}
\end{figure*}

\begin{figure}[!htb]
    \centering
     \begin{subfigure}[t]{0.49\textwidth}
        \centering
        \includegraphics[scale=0.4]{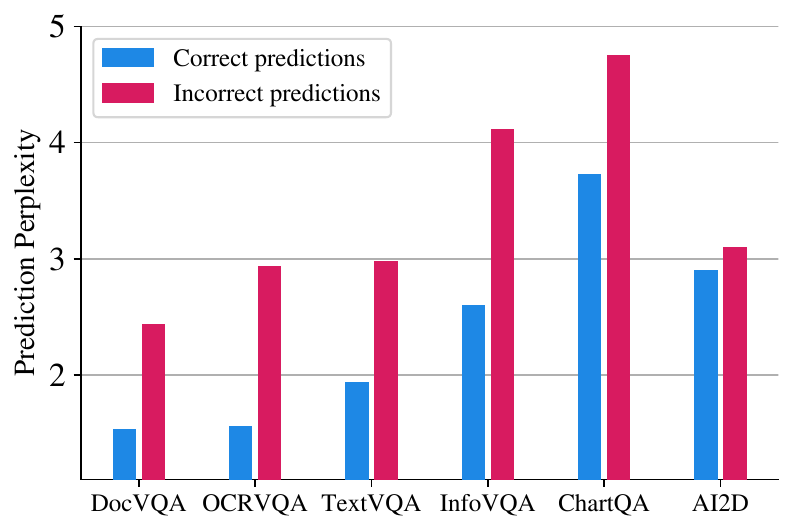}
        \caption{}
         \label{fig:perplexity}
     \end{subfigure}
     \begin{subfigure}[t]{0.49\textwidth}
        \centering
        \includegraphics[scale=0.4]{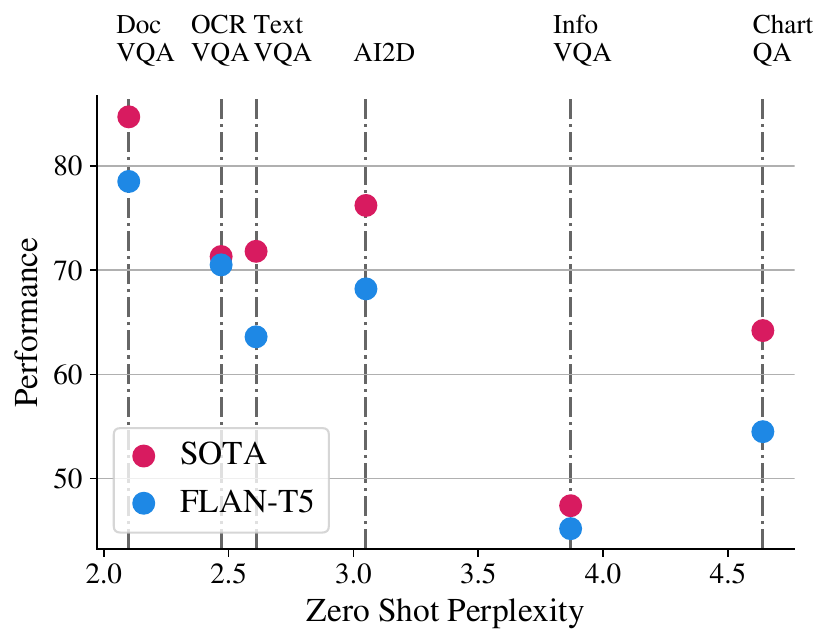}
        \caption{}
        \label{fig:perplexity_vs_score}
     \end{subfigure}
    \caption{(a) \textbf{Perplexity} of zero shot prediction for correct and incorrect sets across all datasets. (b) SOTA and full finetuned performance with Zero Shot Perplexity of FlanT5-B. }
    \label{fig:shuffled-vs-length}
\end{figure}
\balance
{\flushleft \textbf{Definition:}} For a language model $M$, we define reading order perplexity as the perplexity of the predicted tokens given the context, i.e. the reading order text passed to the model for question answering. Formally, given a language model $M$, context $C$ and a sequence of predicted tokens $(p_1,p_2 ...., p_n)$, we define the Reading Order Perplexity ($ROP$) as:

$$ROP_M = \exp\Big(- \frac{1}{n} \sum_{i=1}^n \log P_M(p_i|p_{1:i-1}, C)\Big) $$

where $P_M(p_i|p_{1:i-1}, C)$ represents the likelihood of the $i^{th}$ token predicted by the model $M$.
\label{defn:rop}
Based on Definition \ref{defn:rop}, we define \textit{Zero Shot Perplexity} as the average reading order perplexity of prediction for the dataset in a zero shot setting, i.e., with a pre-trained model.


We calculate the mean perplexity of two sets - correct and incorrect predictions and plot that in Figure \ref{fig:perplexity} for all the datasets. As evident from the chart, zero shot perplexity correlates strongly with the model performance, with significantly lower values for correct predictions compared to incorrect predictions. Also, we observe a strong correlation between zero shot perplexity and the fine-tuned model's performance, as shown in Figure \ref{fig:perplexity_vs_score}. Interestingly, the SOTA performance also seems to show a strong correlation with Zero Shot Perplexity, with lower values of perplexity corresponding to better scores overall. This seems to indicate that even for visual language models which include image and layout features performance, the key to unlocking better results might be through stronger language models. Also, we believe zero shot perplexity also serves as an easy shortcut to estimating model potential and overall performance for unseen tasks.

\paragraph{Key Observations.} The type of reasoning needed to answer image document question answering tasks is one of five types (as shown in Figure \ref{fig:exemplars}) - (a) direct extraction from the content of the image, (b) reasoning based on the text content of the image along with world knowledge, (c) answering solely based on the visual content of the image, (d) answering with no content of the image but only world knowledge, (e) reasoning based on visual as well as text content of the image (layout understanding falls under this category). An LLM only model, as employed here will be able to address question types (a), (b) and (d) exclusively.  The notably strong performance across various datasets implies that a majority chunk of questions fall within these categories which do not even need visual reasoning. For these questions, scaling the language model appears more effective than scaling the vision encoder.
In order to judge whether a vision encoder is able to extract the useful features necessary to answer questions, we should be looking at the genre of question-answers in (d) and (e). Hence, the reading order based language only scheme used in this paper can serve as a question segregator to create subsets of datasets on which the improvements of a new vision encoder should be shown. 

\section{Conclusion}
In this paper, we analyze the contribution of an LLM in image document question answering tasks, by modeling it as an LLM-only task, without any vision encoder. Specifically, we show that sending the text present in an image as an ordered set of tokens to an LLM, we can achieve near-SOTA performance on a variety of  benchmark datasets of visual question answering which involve text in image. We analyze multiple factors which play important roles towards the effectiveness of such an LLM-only model. We hope that this can guide the choice of models for practitioners, and choice of datasets for researchers developing models for which a vision encoder is truly essential. 


\clearpage
{
\bibliographystyle{ieee_fullname}
\bibliography{refs}
}
\appendix
\clearpage

\section{Multi-Tasking}
Prior works such as T5 \cite{t5} have observed that multi-tasking during pre-training and fine-tuning can boost generalization capabilities through transfer of abilities across tasks. To explore this, we experiment with training a single model on multiple tasks. We finetune FlanT5-B and FlanT5-XXL by creating a multi-task mixture of these four datasets - DocVQA, OCRVQA, InfoVQA and TextVQA. As it is intuitive, the mixing proportions seem to play an important role in multi-tasking. In this work, we experiment with two mixing strategies :
\begin{itemize}
    \item \textbf{Uniform}: One of the datasets is first chosen uniformly with equal probability and then one sample is chosen randomly from that dataset.
    \item \textbf{Normalized}: In this we pool all the datasets together and then randomly sample an example from it.
\end{itemize}
The results are shown in Table \ref{table:multitask}. For FlanT5-B, the two settings perform closely. However, note that the uniform strategy performs slightly better for most datasets, except OCR-VQA.  This is because in the normalized setting, the model sees more data points from OCR-VQA, which may lead to allotting more weight to this task. For FlanT5-XXL, we observe the reverse trend. Here, the normalized setting consistently performs better by a significant margin. In some cases like OCR-VQA, DocVQA, the mixture setting shows further improvement over the best reported results in the paper. This suggests that multi-tasking may be useful for learning relevant features across different datasets.

\begin{table}[h]
  \caption{\textbf{Multi-Task Finetuned} performance for document QA with reading order input}
  \label{table:multitask}
  \centering
  \resizebox{0.47\textwidth}{!}{
    \begin{tabular}{@{}lccccc@{}}
    \toprule
    Model            & Setting & OCR-VQA & DocVQA & InfoVQA & TextVQA \\ \midrule
    \multirow{2}{*}{FlanT5-B}   & Uniform     &   69.0      &  68.7      &     33.5           &    56.2     \\
             &  Normalized  &     69.6  &   68.3     &    31.5       &        55.5   \\
    \multirow{2}{*}{FlanT5-XXL}   & Uniform      & 70.4 & 72.8 & 41.9 &    58.2  \\
    & Normalized   &    \textbf{72.9}    &    \textbf{78.6}    &   44.9      & 63.5   \\
    \midrule
     Single Task (XXL)       &   - &  70.5    &    78.5     &         \textbf{45.2}      &     \textbf{63.6}   \\
    \midrule
    {\color{gray}
    \multirow{2}{*}{SOTA }} 
     & & {\color{gray} 71.3} & {\color{gray} 84.7} & {\color{gray} 47.4} & {\color{gray} 71.8}\\
     & & {\color{gray} \tiny (Pix2Struct)} & {\color{gray} \tiny (UDOP)} & {\color{gray} \tiny (UDOP)} & {\color{gray} \tiny (PaLI)} \\

    \bottomrule
    \end{tabular}
    }
\end{table}



\section{Test Set Results}
In Table \ref{table:test_set}, we report the test set performance for our best performing model on all the datasets. The performances are obtained from the evaluation servers of the respective datasets. Note that for ChartQA and AI2D, we already report the test set scores in the main paper.
\begin{table}[h]

  \caption{Test Set results on different datasets}
  \label{table:test_set}
  \centering
  \resizebox{0.47\textwidth}{!}{
    \begin{tabular}{@{}clccccccc@{}}
    \toprule
    &            &  OCR-VQA & DocVQA & InfoVQA & TextVQA & ChartQA & AI2D \\ 
    \midrule
    & FLanT5-XXL     &     70.6    &    74.2     &         41.9      &     58.5  & 54.2 & 68.2 \\
    \bottomrule
    \vspace{-10ex}
    \end{tabular}
    }
\end{table}

\section{Dataset-wise analysis}
Here we analyze each dataset based on five broad categories of question-answer pairs as discussed in the main paper - (a) direct extraction from the content of the image, (b) reasoning based on the text content of the image along with world knowledge, (c) answering solely based on the visual content of the image, (d) answering with no content of the image but only world knowledge, (e) reasoning based on visual as well as text content of the image (layout understanding falls under this category).

\begin{figure*}[!htb]
    \centering
     \begin{subfigure}[t]{0.85\textwidth}
            \centering
           \includegraphics[scale=0.32]{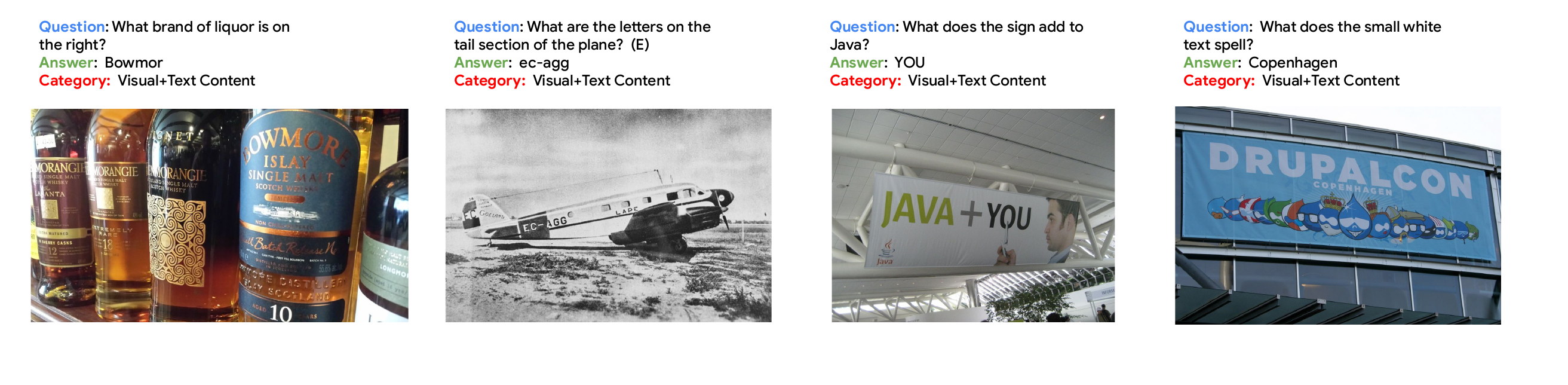}
           \vspace{-10ex}
            \caption{TextVQA}
            \label{fig:exemplars_textvqa}
    \end{subfigure}
    \vfill
        \begin{subfigure}[t]{0.85\textwidth}
            \centering
            \includegraphics[scale=0.32]{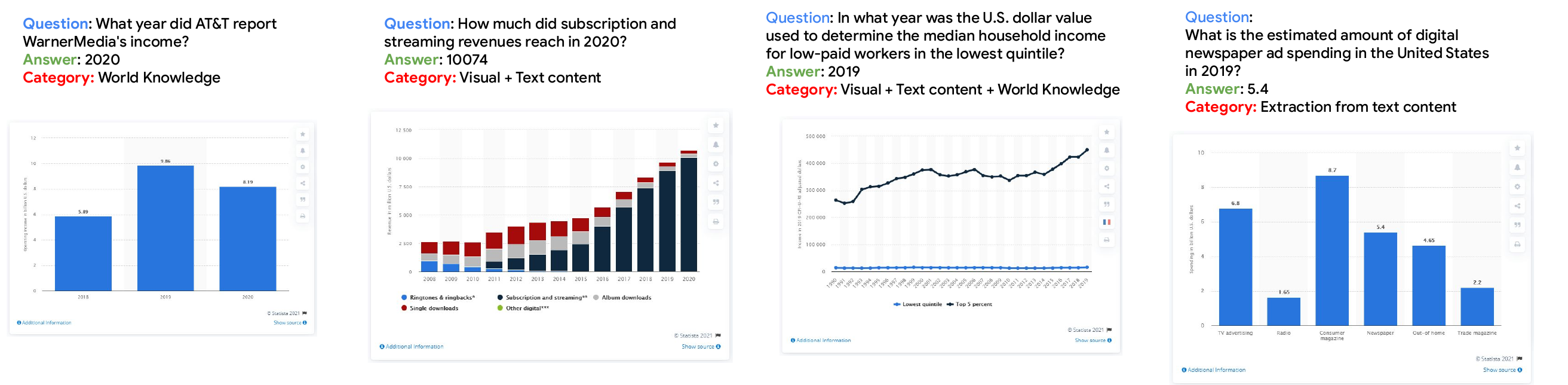}
            \caption{ChartQA}
            \label{fig:exemplars_chartqa}
    \end{subfigure}
    \vfill
            \begin{subfigure}[t]{0.85\textwidth}
            \centering
            \includegraphics[scale=0.32]{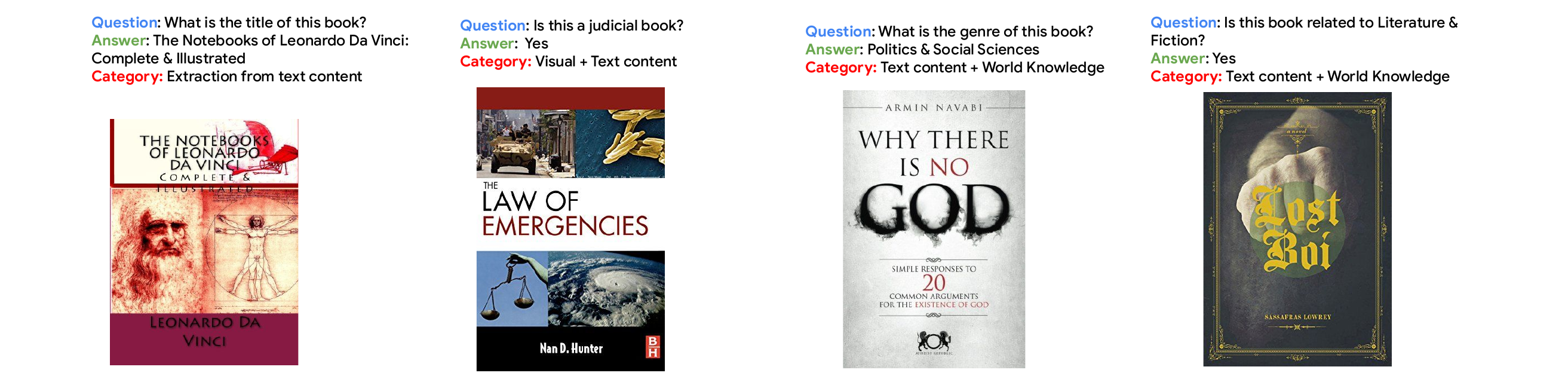}
            \caption{OCR-VQA}
            \label{fig:exemplars_ocrvqa}
    \end{subfigure}
                \begin{subfigure}[t]{0.85\textwidth}
            \centering
            \includegraphics[scale=0.32]{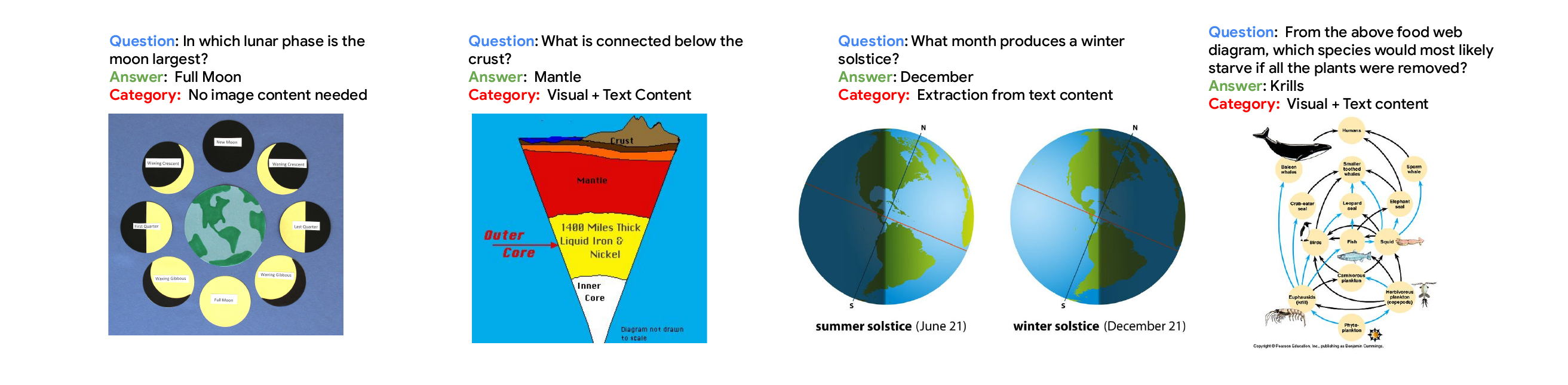}
            \caption{AI2D}
            \label{fig:exemplars_ai2d}
    \end{subfigure}
            \begin{subfigure}[t]{0.85\textwidth}
            \centering
            \includegraphics[scale=0.32]{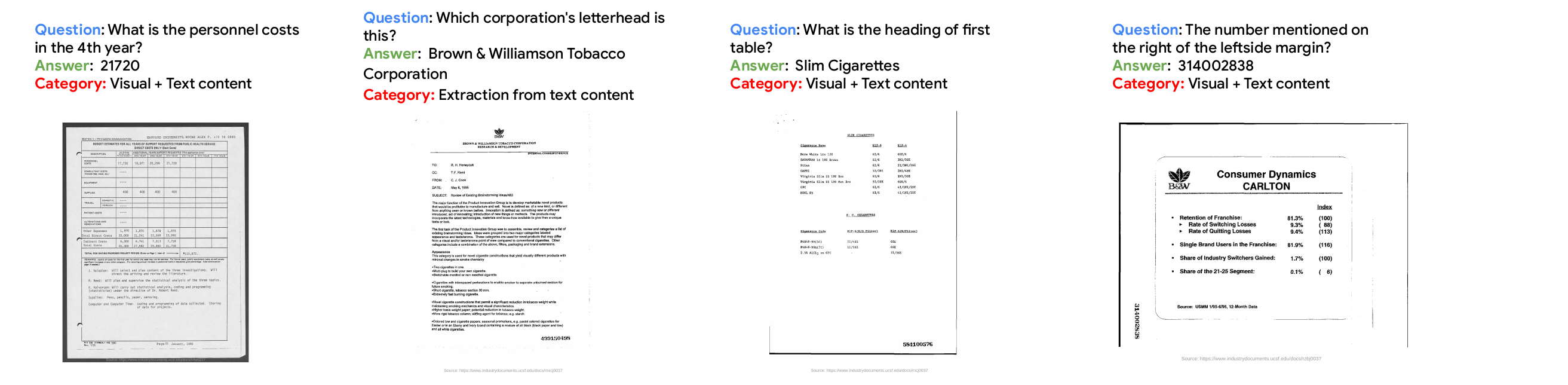}
            \caption{DocVQA}
            \label{fig:exemplars_docvqa}
    \end{subfigure}
\vspace{-5ex}
        \begin{subfigure}[t]{0.85\textwidth}
            \centering
            \includegraphics[scale=0.32]{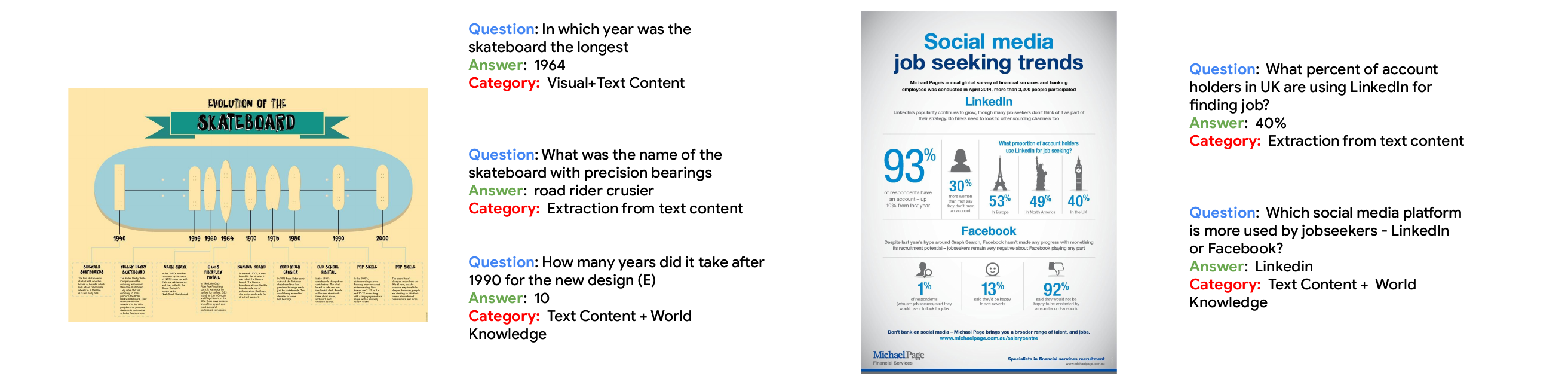}
            \caption{InfoVQA}
            \label{fig:exemplars_infovqa}
    \end{subfigure}
    \label{fig:supp_exemplars}
\end{figure*}

\textbf{TextVQA}: This dataset has a lot of questions which fall into category (e) as they need some visual understanding of the image along with the text content and the world knowledge corresponding to the visual entities. Figure \ref{fig:exemplars_textvqa} represents such cases. In the four cases highlighted, the model needs a visual understanding of the relative positions, object structure, symbols and font sizes respectively to infer the answer, in addition to the OCR text. 

\textbf{ChartQA}: Questions in this dataset fall into category (a) and (e). The answer can either be extracted from the OCR-text directly or the model needs a visual understanding of the charts. We also observe some questions which cannot be answered solely by the information present in the graph. Such examples fall into category (d). Figure \ref{fig:exemplars_chartqa} shows examples from this dataset.

\textbf{OCRVQA}: Questions in this dataset fall into category (a), (b) and (e) (Figure \ref{fig:exemplars_ocrvqa}). In most cases, the model can infer the answer based on the OCR text content, along with some domain knowledge from the world. We believe that this is one of the easy datasets for a setting where visual encoder is absent. Our empirical findings are consistent with this hypothesis, as we are able to perform on par or even better than SOTA models (refer Table \ref{table:multitask}) using our LLM-only approach.

\textbf{AI2D} (\ref{fig:exemplars_ai2d}): This is a vision-heavy dataset but we find that most questions can still be answered with OCR context and enough world knowledge, hence falling into category (b) and (d) along with a few cases from category (e).

\textbf{DocVQA} (\ref{fig:exemplars_docvqa}): This is a text-heavy dataset. While both the OCR text content and the layout information are required to accurately answer a given query,  we still observe a decent performance, without using any layout information, with our \textit{Raster Scan} reading order strategy. Questions in this dataset fall either into category (a) or (e).

\textbf{InfoVQA} (\ref{fig:exemplars_infovqa}): This dataset mostly consists of instances from category (a), (b) and (e). Usually, some world knowledge is required to infer the correct answer from the text component of the given infographic. Along with this, we also observe cases where the visual information is critical to obtain the right response (1st case in Figure \ref{fig:exemplars_infovqa}).


\end{document}